\documentclass{article}

\usepackage[final]{corl_2017} % Uncomment for the camera-ready ``final'' version

\usepackage{graphics} % for pdf, bitmapped graphics files
\usepackage{epsfig} % for postscript graphics files
\usepackage{mathptmx} % assumes new font selection scheme installed
\usepackage{times} % assumes new font selection scheme installed
\usepackage{amsmath} % assumes amsmath package installed
\usepackage{amssymb}  % assumes amsmath package installed

\usepackage{subfigure}
\newcommand\tab[1][0.5cm]{\hspace*{#1}}
\usepackage{wrapfig}

\title{Learning a visuomotor controller for real world robotic grasping using simulated depth images}

\author{
  Ulrich Viereck\textsuperscript{1}, Andreas ten Pas\textsuperscript{1}, Kate Saenko\textsuperscript{2}, Robert Platt\textsuperscript{1}\\ \\
  \textsuperscript{1}College of Computer and Information Science, Northeastern University\\
  360 Huntington Ave, Boston, MA 02115, USA\\
  \texttt{\{uliv,atp,rplatt\}@ccs.neu.edu} \\ \\
  \textsuperscript{2}Department of Computer Science, Boston University\\
  111 Cummington Mall, Boston, MA 02215\\
  \texttt{saenko@bu.edu} \\
}

\begin{document}
\maketitle

%===============================================================================
\vspace{-0.2in}
\begin{abstract}
We want to build robots that are useful in unstructured real world applications, such as doing work in the household. Grasping in particular is an important skill in this domain, yet it remains a challenge. One of the key hurdles is handling unexpected changes or motion in the objects being grasped and kinematic noise or other errors in the robot. This paper proposes an approach to learning a closed-loop controller for robotic grasping that dynamically guides the gripper to the object. We use a wrist-mounted sensor to acquire depth images in front of the gripper and train a convolutional neural network to learn a distance function to true grasps for grasp configurations over an image. The training sensor data is generated in simulation, a major advantage over previous work that uses real robot experience, which is costly to obtain. Despite being trained in simulation, our approach works well on real noisy sensor images. We compare our controller in simulated and real robot experiments to a strong baseline for grasp pose detection, and find that our approach significantly outperforms the baseline in the presence of kinematic noise, perceptual errors and disturbances of the object during grasping.

\end{abstract}

\keywords{Robots, Learning, Manipulation}

%%%%%%%%%%%%%%%%%%%%%%%%%%%%%%%%%%%%%%%%%%%%%%%%%%%%%%%%%%%%%%%%%%%%%%%%%%%%%%%%
\section{Introduction}\label{Introduction}
%\vspace{-0.05in}
Recently, deep neural networks have been used to learn a variety of visuomotor skills for robotic manipulation including grasping, screwing a top on a bottle, mating a mega-block, and hanging a loop of rope on a hook~\cite{Levine_jmlr_2016}. Grasping is a particularly useful and ubiquitous robotics task. A number of researchers have recently proposed using deep learning for robotic grasping systems that perform well for novel objects presented in dense clutter~\cite{Gualtieri2016,Lenz2015,Pinto2016}. However, these systems still do not perform as well as we would like, achieving maximum grasp success rates of approximately 85\% to 93\% in ideal conditions~\cite{Gualtieri2016}. The question is how to learn robotic grasping or manipulation behaviors that are robust to the perceptual noise, object movement, and kinematic inaccuracies that occur in realistic conditions.

\newcommand{\widthb}{1.2in}
\begin{figure}
    \centering
    \hspace{-0.07in}
    \subfigure[]{\includegraphics[width=\widthb]{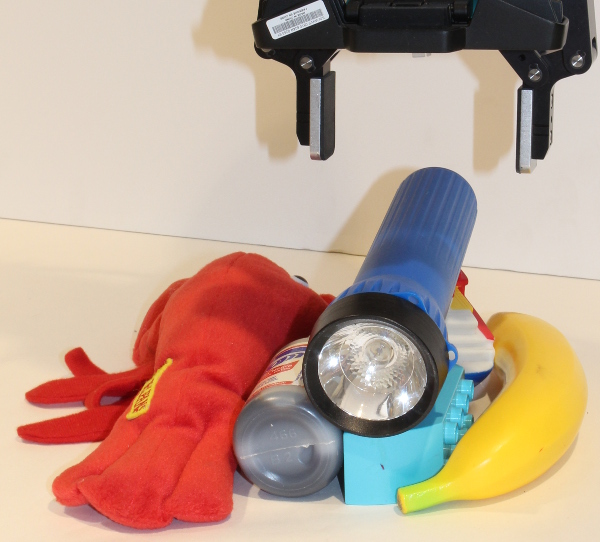}}
    \subfigure[]{\includegraphics[width=\widthb]{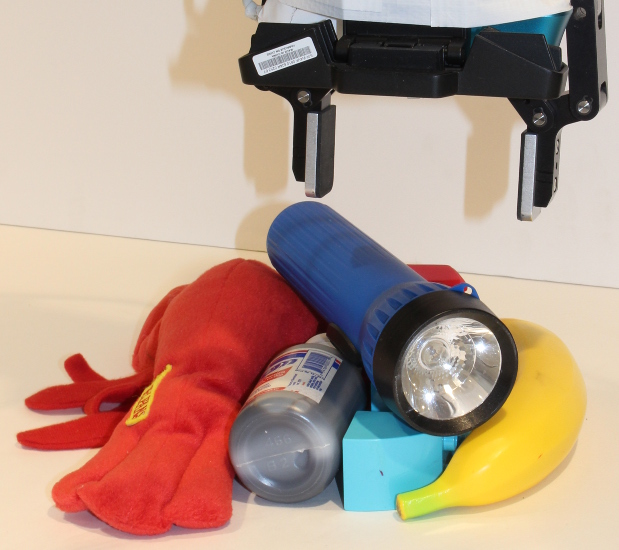}}
    \subfigure[]{\includegraphics[width=\widthb]{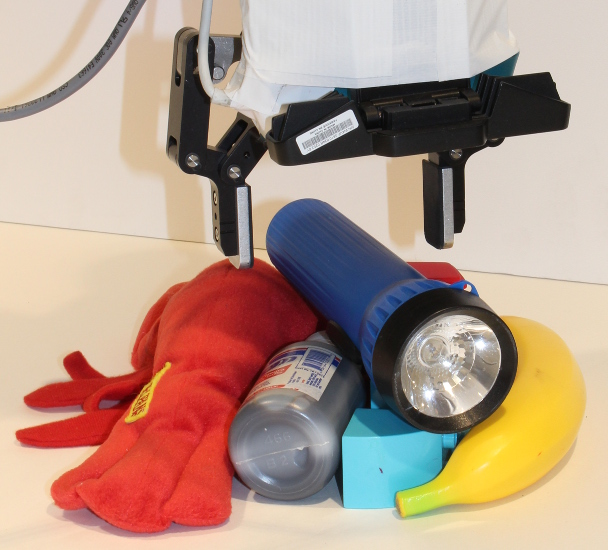}}\\ \vspace{-0.1in}
    \subfigure[]{\includegraphics[width=\widthb]{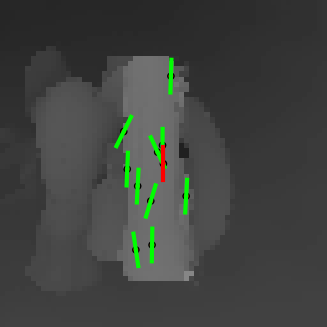}}
    \subfigure[]{\includegraphics[width=\widthb]{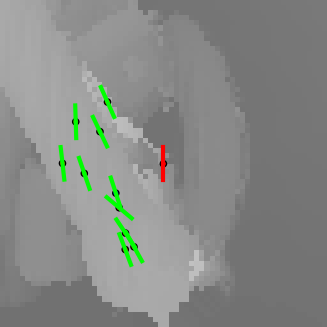}} 
    \subfigure[]{\includegraphics[width=\widthb]{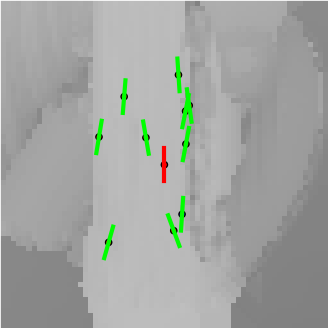}} \vspace{-0.1in}    
    \caption{Our controller makes dynamic corrections while grasping using depth image feedback from a sensor mounted to the robot's wrist. (a) The hand has moved to the initial detected grasping position for the flashlight. (b) The flashlight has shifted and the hand became misaligned with the object. (c) The controller has corrected for the misalignment and has moved the hand into a good grasp pose. The controller is now ready to pick up the flashlight. (d) - (f) show the corresponding depth image. The green lines show initial grasps predicted by the CNN. The red line shows the current gripper pose.}
    \label{gripper_with_clutter}
    \vspace{-0.17in}
\end{figure}

A major problem with many existing approaches is that they perform one-shot grasp detection and thus cannot learn dynamic correcting behaviors that respond to changes in the environment. One promising solution is to learn a closed-loop visuomotor controller. In contrast to one-shot grasp detection, closed-loop controllers have the potential to react to the unexpected disturbances of the object during grasping that often cause grasps to fail. The recent work by Levine et al.~\cite{Levine2016} used supervised deep networks to learn a closed-loop control policy for grasping novel objects in clutter. However, their approach has two important drawbacks. First, it requires visual data that observes the scene from a specific viewpoint with respect to the robot and the scene. The consequence of this is that it is difficult to adapt the learned controller to a different grasping scene, e.g., a different table height or orientation relative to the robot. Second, their approach requires two months of real world training experience. In many scenarios, it is simply not practical to obtain such a large quantity of robotic training data.

This paper proposes an approach to closed-loop control for robotic manipulation that is not subject to either of the two limitations described above. We make three key contributions. First, in order to eliminate the dependence on a particular viewing direction, we mount a depth sensor near the robot end-effector as shown in Figure~\ref{gripper_with_clutter}. In this configuration, the same visuomotor controller can be used to grasp objects from any direction, because the camera to gripper configuration is fixed. Second, we train the system completely in simulation, thereby eliminating the dependence on enormous amounts of real-world robotic training experience. The key to training in simulation is our use of depth sensors rather than RGB cameras. While depth data is potentially less informative than RGB data, it can be simulated relatively accurately using ray tracing (we use OpenRAVE~\cite{Diankov2008}). Third, we propose a novel neural network model that learns a distance-to-nearest-grasp function used by our controller. Our convolutional neural network (CNN) is similar in structure to that of Levine et al.~\cite{Levine2016}, but takes images at a lower resolution and has many fewer layers. Instead of learning a policy directly, we learn a distance function, i.e., distance to grasp, using CNN regression with an L1 loss function. This function provides direct feedback about how viable a grasp is and allows us to use a simple controller to move the robot arm. We evaluate the performance of the system both in simulation and on a UR5 robot in our lab. Our major finding is that in the absence of motor or sensor noise, our closed-loop grasp controller has similar performance to a recently developed grasp detection method~\cite{Gualtieri2016} with very high grasp success rates. However, under realistic motor, kinematic and sensor errors, the controller proposed here outperforms that method significantly.

%%%%%%%%%%%%%%%%%%%%%%%%%%%%%%%%%%%%%%%%%%%%%%%%%%%%%%%%%%%%%%%%%%%%%%%%%%%%%%%%
\section{Related Work}
\vspace{-0.05in}
Recent work in grasp perception has utilized deep learning to localize grasp configurations in a way that is analogous to object detection in computer vision~\cite{Lenz2015,Kappler2015,Redmon2015,Pinto2016,Mahler17}. Such methods take potentially noisy sensor data as input and produce viable grasp pose estimates as output. 
However, these grasp detection methods typically suffer from perceptual errors and inaccurate robot kinematics~\cite{Gualtieri2016}. In addition, extending traditional one-shot grasp perception methods to re-detect grasps in a loop while the sensor mounted on the gripper gets closer to the objects is difficult, because these approaches are trained to find grasps with large distances to the sensor (e.g., to see the entire object)~\cite{Gualtieri2016,Lenz2015,Pinto2016,Redmon2015}.

Visual servoing methods use visual feedback to move a camera to a target pose that depends directly on the object pose. While there are numerous methods in this area~\cite{Siciliano2007}, only a small amount of previous work addresses using visual feedback directly for grasping~\cite{Vahrenkamp2008,Hebert2012,Kapach2012}. In contrast to our work, the existing methods require manual feature design or specification. An active vision approach by Arruda et al. acquires sensor data from different view points to optimize surface reconstruction for reliable grasping during grasp planning~\cite{Arruda16}. However, the actual grasping does not use sensor feedback. 

Levine et al. were one of the first to incorporate deep learning for grasp perception using visual feedback~\cite{Levine2016}. However, their approach requires months of training on multiple physical robots. Moreover, they require a CNN with 17 layers that must be trained from scratch. In addition, their use of a static camera makes it difficult to adapt to different grasping scenarios, e.g., a different table height or a different grasp approach direction. Because we generate training data in simulation and our CNN has only a few layers, our approach is simpler. In addition, since we mount the camera to the wrist of the robot arm, our approach is more flexible because it can be applied to any grasping scenario -- not just those with a particular configuration relative to the camera.

%%%%%%%%%%%%%%%%%%%%%%%%%%%%%%%%%%%%%%%%%%%%%%%%%%%%%%%%%%%%%%%%%%%%%%%%%%%%%%%%
\section{Approach}
\vspace{-0.05in}
\begin{figure*}[t]
      \centering
      \includegraphics[width=1.0\textwidth]{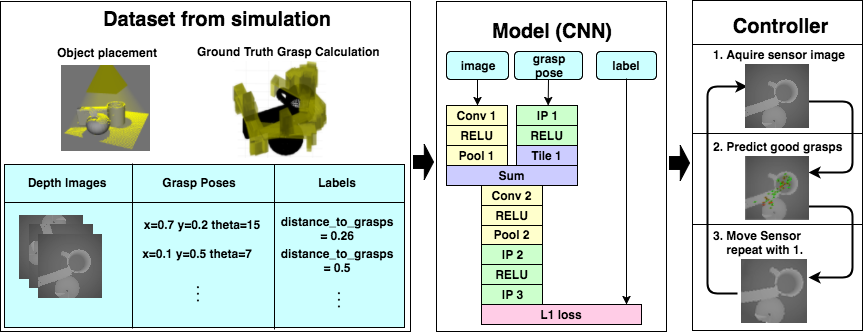}
      \caption{Overview of our approach. The training data is generated in an OpenRAVE simulator (\ref{dataset}). A CNN model is trained to predict distance to nearest grasps (\ref{model}). A controller moves the gripper to predicted good grasp poses (\ref{controller}).
      }
      \label{overview}
      \vspace{-0.1in}
\end{figure*}

We propose a new approach to the problem of learning a visuomotor controller for robotic grasping inspired by the method of Levine et al.~\cite{Levine2016}. We mount a depth sensor near the wrist of the robot as shown in Figure~\ref{gripper_with_clutter}. On each control step, the system takes a depth image of the scene directly in front of the gripper and uses this sensor information to guide the hand. The controller converges to good grasp configurations from which the gripper fingers can close and pick up the object. The approach is based on a convolutional neural network that learns a distance function. It takes the depth image in conjunction with a candidate hand displacement as input and produces as output an estimate of the distance-to-nearest-grasp. Figure~\ref{overview} shows an overview of the approach. The key elements are: 1) the convolutional neural network that is used to model the distance function (Section \ref{model}); 2) the approach to generating the training set in simulation (Section \ref{dataset}); 3) the implementation of the controller (Section \ref{controller}).

%%%%%%%%%%%%%%%%%%%%
\subsection{CNN Model}\label{model}
\vspace{-0.05in}
The core of our work is a convolutional neural network (a CNN, see Figure~\ref{overview}) that learns a distance function that is used by our grasp controller. The network takes as input a depth image, $I$, and an action, $a = (x,y,\theta) \in \mathbb{R}^2 \times \mathbb{S}^1$. The action denotes a candidate planar pose offset relative to the depth sensor to which the robotic hand could be moved. It learns a real-valued function, $d(I,a) \in \mathbb{R}_{>0}$, that describes the distance between the hand and the nearest viable grasp after displacing the hand by $a$. We interpret this distance to be the remaining cost-to-go of moving to the nearest viable grasp after executing action $a$. Distance is measured in meters in the $(x,y,\theta)$ pose space by weighting the angular component (by 0.001 meter/degree) relative to the translational parts.

Our CNN is based on the LeNet network designed for handwritten digit classification~\cite{Lecun1998}. It consists of two convolutional layers (Conv1 with 20 and Conv2 with 50 filters, kernel size 5, and stride 1) with leaky RELUs, max pooling and 2 inner-product (IP) layers with leaky RELUs. Inspired by Levine et al.~\cite{Levine2016} we apply an IP layer to the input pose vector (action) and then tile the resulting output over the spatial dimensions to match the dimensions of the Pool1 layer and sum element-wise. The output layer predicts the distance-to-go for the grasp pose action. Since we are learning a real-valued distance function, our CNN is solving a regression problem. We also tried a classification model, but we found that the controller (Section \ref{controller}) using the regression model performs better because the predictions from the regression allows to compare the goodness of two grasp poses, where the better pose is \emph{closer} to a true grasp. We evaluated both L1 and L2 loss functions and found the L1 loss function did a better job fitting our data. 

%%%%%%%%%%%%%%%%%%%%
\subsection{Generating training data}\label{dataset}

\begin{wrapfigure}{r}{0.5\textwidth}
      \vspace{-0.15in}
      \centering
      \includegraphics[width=0.5\textwidth]{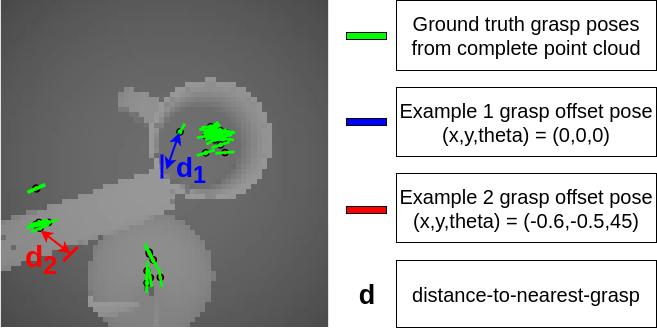}
      \caption{Calculating the distance-to-nearest-grasp for two different offset poses (shown in red and blue). During creation of the training set, we estimate the distance between each of these pose offsets and the nearest ground truth grasp (shown in green).}
      \label{input_grasp_pose_with_distance}
      \vspace{-0.1in}
\end{wrapfigure}

We create a dataset in simulation using OpenRAVE~\cite{Diankov2008} comprised of image-offset pairs and the corresponding distance-to-nearest-grasp labels. The way that OpenRAVE simulates the depth images is of particular interest. If the simulated images are sufficiently different from the images generated by an actual depth sensor, then this would produce a gap that would make it difficult to transfer the learned policies onto the real robot. % There is still ``policy" here
Fortunately, we found that this was not the case. The model learned on depth images generated by OpenRAVE (using ray tracing) seems to transfer well (Figure~\ref{real_results}).

\begin{wrapfigure}{r}{0.5\textwidth}
%\begin{figure}[thpb]
      \vspace{-0.18in}
      \centering
      \includegraphics[width=0.5\textwidth]{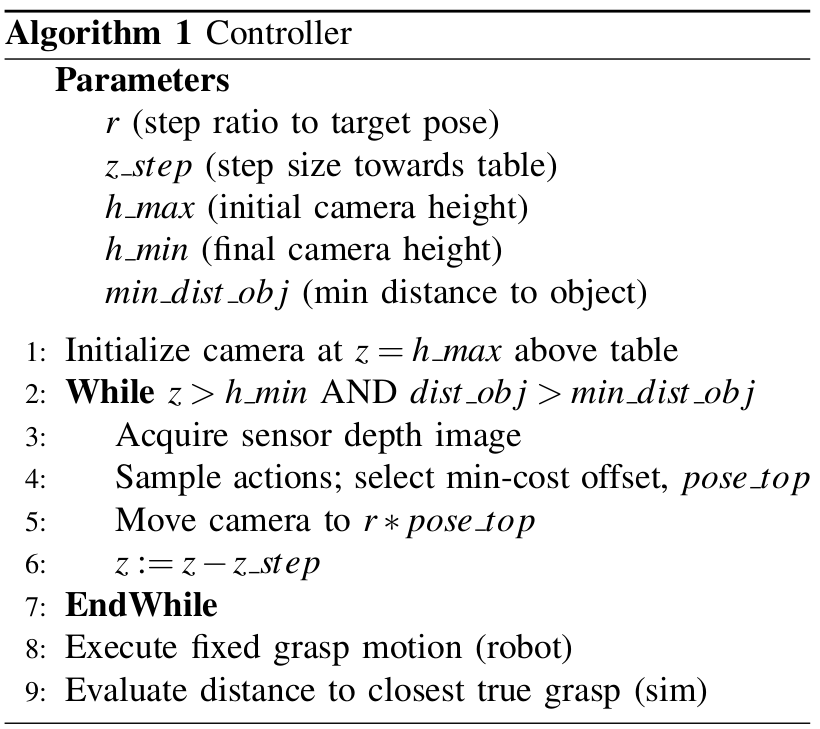}
      \vspace{-0.2in}
%\end{figure}
\end{wrapfigure}

In order to train the CNN, we generate a large number of image-action pairs, each associated with a distance-to-nearest-grasp label. We accomplish this using OpenRAVE as follows. First, we generate 12.5k different scenes with a random selection of multiple objects placed under the sensor. The objects were derived from CAD models contained within the 3DNet database~\cite{Wohlkinger2012}. In particular, we have selected 381 graspable objects from the following 10 categories: mug, hammer, bottle, tetra pak, flash light, camera, can, apple and toy car. There are between 1-5 CAD objects in each scene (the number of objects is uniformly sampled). Each object is placed with a random position and orientation. Figure~\ref{input_grasp_pose_with_distance} shows a depth image with a cup, apple and camera.

For each scene we generate 40 depth images by placing the camera randomly in ($x,y,z,\theta$) above the objects, where $x,y$ are the directions parallel to the table and $z$ is the direction towards the table. This results in a total of 500k depth images. Each depth image has one color channel (grayscale) and has a size of $64 \times 64$ pixels. For each depth image we uniformly sample 10 offset poses within the camera view and calculate the distance to the nearest grasp for each pose as follows. First, using the mesh model of the scene, we sample a large number of grasp candidates by filtering for robotic hand poses that are collision free and that contain parts of the visible environment between the robotic fingers (see~\cite{tenpas_isrr2015}). Then, we test each candidate for force closure using standard methods~\cite{Nguyen86}. Finally, after pruning the non-grasps, we evaluate the Euclidean distance to the nearest sampled grasp (see Figure~\ref{input_grasp_pose_with_distance}).

%%%%%%%%%%%%%%%%%%%%
\subsection{Controller}\label{controller}

Our controller takes actions that descend the distance function that is modelled by the CNN described in Section~\ref{model}. Its basic operation is outlined in Algorithm~1. The controller starts with the hand at a fixed initial height above the table in the $z$-direction. In Step 3, the controller acquires an image from the wrist-mounted depth sensor. In Step 4, it samples a set of candidate actions and selects the one with the minimum distance-to-nearest-grasp. In Step 5, the controller moves by a constant fractional step size in the direction of the selected action. The fact that the controller only makes a fractional motion on each time step smooths the motion and makes the controller more robust to isolated bad predictions by the CNN. In Step 6, the controller approaches the object in the $z$-direction by one step. This process repeats until the controller converges and the hand reaches the final hand height.

\begin{wrapfigure}{r}{0.5\textwidth}
      \vspace{-0.15in}
      \centering
      \includegraphics[width=0.5\textwidth]{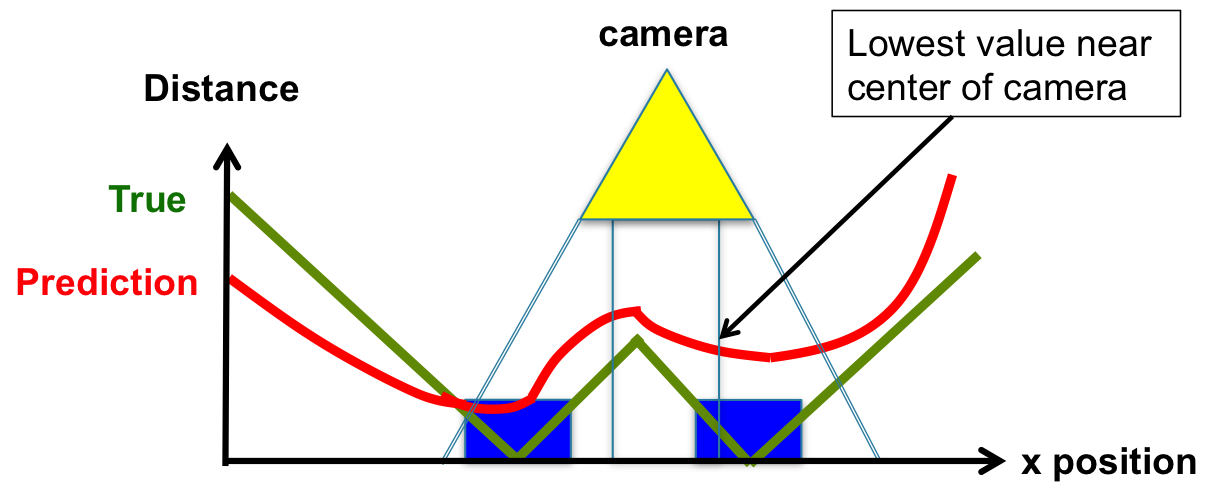}
      \caption{Illustration for how the controller works in a 1-dimensional case with two objects (control in x-axis direction). Although the global best prediction for the grasp pose belongs to the object on the left, the controller moves to the \emph{closer} object on the right, because it follows the direction of the local gradient near the center of the image.}
\label{fig_controller}
\end{wrapfigure}

An important point is that we constrain the sampling to a region around the origin. This serves two purposes. First, it reduces the number of samples needed. Second, it enables us to capture the gradient of the distance function in the neighborhood of the current hand pose. The distance function may be a multimodal function in the space of offsets. For purposes of stability, it is important for the controller to follow a gradient. In our case, that gradient is estimated in the neighborhood of the center of the image. This is illustrated in Figure~\ref{fig_controller}. Although the global minimum in the distance function is on the left, the hand will follow the gradient to the right. 
The controller thus grasps the object closest to the current hand pose, regardless of its identity. If our goal was to grasp a desired target object, our approach could be extended to first run object detection and then sample grasp candidates near the target object, e.g., within a bounding box around it.

%%%%%%%%%%%%%%%%%%%%%%%%%%%%%%%%%%%%%%%%%%%%%%%%%%%%%%%%%%%%%%%%%%%%%%%%%%%%%%%%
\section{Simulation Experiments}\label{sim_results}
\vspace{-0.05in}
We perform a series of experiments in simulation to evaluate our new grasp controller (CTR) relative to grasp pose detection (GPD), a recently proposed one-shot method that also learns in simulation and achieves high success rates~\cite{Gualtieri2016}. We perform this comparison for two scenarios: one where the manipulator moves exactly as commanded and one where the desired manipulator motions are corrupted by zero-mean Gaussian noise. All of the following simulation data are averages over 400 trials. In each trial, we generate a scene in OpenRAVE with a random selection and placement of objects from the test set as described in Section~\ref{dataset}. The initial camera position is set to 0.3 m above the table. At each iteration the camera height is reduced by a constant step until height 0.15 m is reached. We run the controller for a total of 75 iterations, using $r=0.2$ as the step ratio to a target pose, and plot the distance of the final gripper pose to the closest true grasp.

We use the deep learning framework Caffe~\cite{Jia2014} for training the network. We run 900k iterations of stochastic gradient descent with a learning rate of 0.001, a momentum of 0.9, and a batch size of 1k instances. The dataset is described in Section~\ref{dataset}. We split the training and test sets on object instances. Both the training and test sets contain all 10 object categories. However, the same object \textit{instance} does not appear in both sets. The training set contained 331 object instances and the test set contained 50. We use the same network for all experiments, including experiments on the robot in Section~\ref{real_results}.

\begin{figure}[thpb]
      \centering
      \includegraphics[scale=0.395]{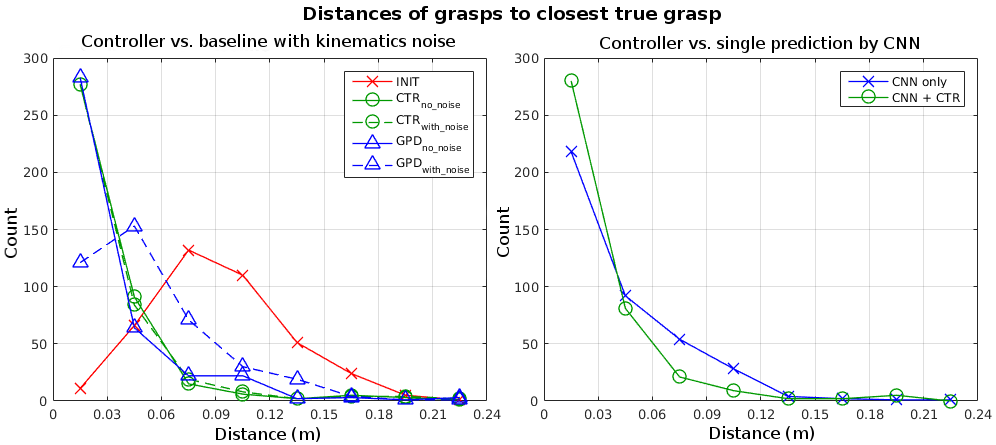}
      \caption{Histogram of distances of predicted grasps to closest true grasp for 400 simulated trials for various scenarios (bin size = 3 cm). Left plot shows that our approach (CTR) compensates well for movement noise of the gripper, where the baseline method (GPD) fails to compensate. Right plot shows that our closed-loop controller compensates for perceptual errors made in the first images by making corrections based on new images while moving to the grasp.}
      \label{hist1_hist2}
      \vspace{-0.15in}
\end{figure}

\subsection{Comparison with GPD baseline in the presence of kinematic noise}

We compare the following scenarios: 1) $INIT$: No camera motion, distances measured from the initial position; 2) $CTR_{no\_noise}$: Run CTR starting from the initial position, without kinematics noise; 3) $CTR_{with\_noise}$: Run CTR from the initial position, with kinematics noise; 4) $GPD_{no\_noise}$: Move to top GPD prediction, without kinematics noise; 5) $GPD_{with\_noise}$: Move to top GPD prediction, with kinematics noise. The ``with\_noise'' scenario, simulates the situation where uncorrelated zero mean Gaussian noise is added to each dimension of the robotic hand displacements on each control step:
\begin{equation}
\Delta (x,y,\theta)_{noisy} = \Delta (x,y,\theta) + 0.4w \|\Delta (x,y,\theta)\| \tab \tab w \sim \mathcal{N}(0,1) \in \mathbb{R}^2 \times \mathbb{S}^1
\end{equation}
% \Delta (x,y,\theta)_{noisy} = \Delta (x,y,\theta) + 0.4w\Delta (x,y,\theta)
While this noisy-motion scenario is not entirely realistic because real manipulator kinematic errors are typically manifested by constant non-linear displacement offsets rather than Gaussian noise, we nevertheless think this is a good test of the resilience of our controller to kinematic errors.

The final distances to the closest true grasp for the 5 scenarios above is shown in Figure~\ref{hist1_hist2} (left). Note that we only consider the distance in ($x,y,\theta$) and not in $z$, because we assume that the distance to the object can be retrieved easily from the depth image. We convert the distances for $\theta$ from degrees to meters as described in~\ref{model}. Notice that without noise, the performance of GPD and CTR is comparable: the two methods move the robotic hand to a grasp pose approximately equally well. However, CTR does much better than GPD in scenarios \textit{with} motion noise. This makes sense because the controller can compensate to some degree for motion errors while GPD cannot. It should also be noted that the distances in Figure~\ref{hist1_hist2} overstate the minimum distances to good grasps. This is because these are distances to the closest \textit{detected} grasp -- not the actual closest grasp, because the method of finding ground truth grasps as described in~\ref{dataset} does not find \textit{all} viable grasps. Nevertheless, the trends in Figure~\ref{hist1_hist2} (left) convey the behavior of the controller.

\subsection{Correction for perceptual errors made in single-shot prediction using the controller}

Next we compare the following two scenarios to characterize the advantages of the closed-loop controller versus one-shot detection. Note that the network is trained with grasp poses globally sampled in the image, not just near the center: 1) $CNN~only$: Move to the top one-shot global prediction using the CNN regression model; 2) $CNN+CTR$: Move to the top one-shot global prediction and then run the controller. Figure~\ref{hist1_hist2} (right) shows that the controller improves the performance of our approach even in a scenario without kinematic noise. This suggests that the controller can compensate for perceptual errors made in a single depth image, and corroborates similar results obtained by Levine et al. in~\cite{Levine2016}.

%%%%%%%%%%%%%%%%%%%%%%%%%%%%%%%%%%%%%%%%%%%%%%%%%%%%%%%%%%%%%%%%%%%%%%%%%%%%%%%%
\section{Robot Experiments}\label{real_results}
\vspace{-0.05in}
We evaluate our grasp controller on the UR5 robot in three experimental scenarios: (i) objects in isolation on a tabletop, (ii) objects in dense clutter, and (iii) objects in dense clutter with a shift in position after a few controller iterations. In these scenarios, we compare our controller (CTR) to grasp pose detection (GPD), a strong baseline~\cite{Gualtieri2016}. A summary of our experimental results is given in Table~\ref{tab:robot_experiments}. The grasp controller is demonstrated in the supplemental video\footnote{\href{http://www.ulrichviereck.com/CoRL2017}{http://www.ulrichviereck.com/CoRL2017}}.

\begin{wrapfigure}{r}{0.5\textwidth}
    \centering
    \vspace{-0.2in}
    \subfigure{\includegraphics[height=0.85in]{setup}}\hspace{0.05in}
    \subfigure{\includegraphics[height=0.85in]{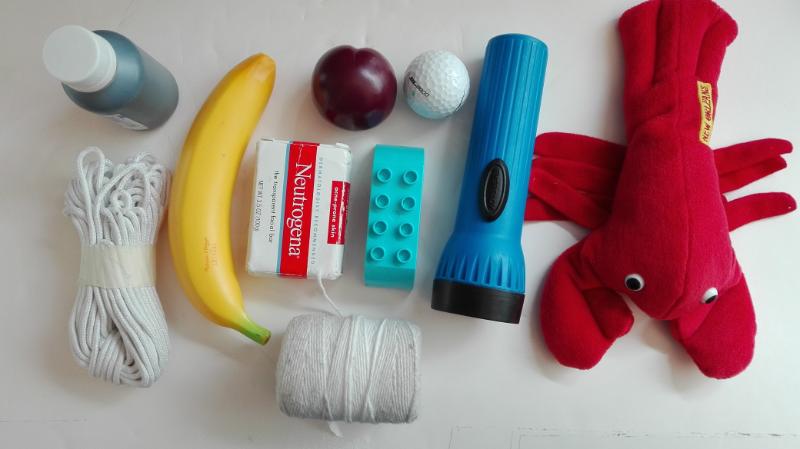}}  
    \vspace{-0.15in}    
    \caption{Experimental setup with UR5 robot and test objects.}
    \label{setup_with_test_objects}
\end{wrapfigure}

We use the UR5, a 6-DOF robot arm, with the Robotiq 85 hand, a 1-DOF parallel jaw gripper with a stroke of 8.5cm, and mount an Intel RealSense SR300 and a Structure IO to the robot's wrist (see Figure~\ref{setup_with_test_objects}). The former sensor is used to obtain depth images for our controller because of its small minimum range (20cm). 
However, point clouds produced by the RealSense are not very accurate and drift with the temperature of the sensor. Because the GPD baseline requires an accurate point cloud, we use the latter sensor. Our controller is implemented in Python on an Intel i7 3.5GHz system (six physical CPU cores) with 32GB of system memory. The control loop runs at about 5Hz. Figure~\ref{setup_with_test_objects} shows the ten objects in our test set. While some of these objects are selected from the same object categories that our CNN model has been trained on (see Section~\ref{dataset}), the specific object instances are not actually contained in the training set.

Each run of our controller proceeds as follows. The arm is first moved to a fixed pose over the table with the depth sensor pointing down toward the table. Then, we run the controller with a step ratio of $r=0.5$ and a $z$ step size of 1cm. Each depth image from the Intel sensor is post-processed to remove invalid depth readings near the edges of objects by applying an iterative, morphological dilation filter that replaces the missing values with a neighborhood maximum. IKFast~\cite{Diankov2008} is used to convert the selected action (i.e., a Cartesian displacement) into target joint angles for the arm to move to. The controller runs until the depth data indicates that an object is within 14~cm distance from the sensor or the robot hand is too close to the table. To execute the grasp, we move the robot hand according to a predefined motion and close the fingers. In total, it takes about 20-30s for each run, depending on the Euclidean distance between the closest object and the robot hand's initial position. The GPD baseline runs as follows. We move the robotic hand to a fixed point above the table pointing directly down and take a depth image using the Structure IO sensor. Then, we select and execute one of the detected grasps based on the heuristics outlined in~\cite{Gualtieri2016}.

\begin{wraptable}[9]{r}{0.5\textwidth}
%\begin{table}[b]
    \centering
    \vspace{-0.15in} 
    \caption{Average grasp success rates for our controller (CTR) and a recent grasp pose detection (GPD) method~\cite{Gualtieri2016} on the UR5 robot.}
    %\begin{scriptsize}
    \vspace{0.1in} 
    \begin{tabular}{|c|c|c|}
        \hline
        \textbf{Scenario} & \textbf{CTR} & \textbf{GPD} \\ \hline
        Objects in isolation & 97.5\% & 97.5\% \\ \hline
        Clutter & 88.9\% & 94.8\% \\ \hline
        Clutter with rotations & 77.3\% & 22.5\% \\ \hline
    \end{tabular}
    \vspace{0.1in}     
    %\end{scriptsize}    
    \label{tab:robot_experiments}
    
%\end{table}
\end{wraptable}

\subsection{Grasping objects in isolation}
In this experiment, each of the ten objects from our test set is presented to the robot in four different configurations: three with the object flat on the table, and the fourth with the object in an upright configuration. Only one out of the 40 grasp attempts results in a failure (97.5\% grasp success rate). This failure happens on the bottle in an upright configuration where our controller is not able to converge to the correct pose. Given more iterations, we would expect the attempt to be successful. In this scenario, GPD achieves the same success rate.

\newcommand{\mywidth}{0.88in}
\begin{figure*}[b]
    \centering
    \vspace{-0.15in} 
    \subfigure{\includegraphics[width=\mywidth]{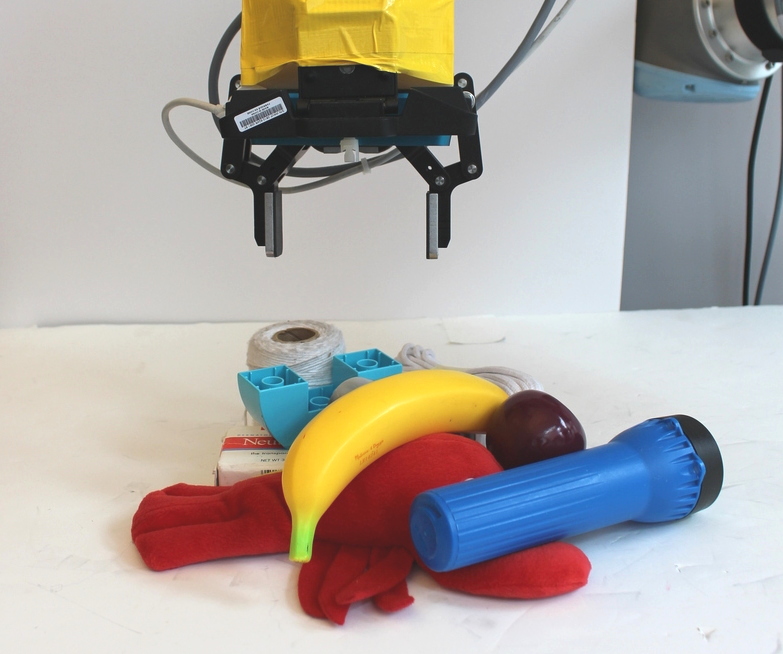}}
    \subfigure{\includegraphics[width=\mywidth]{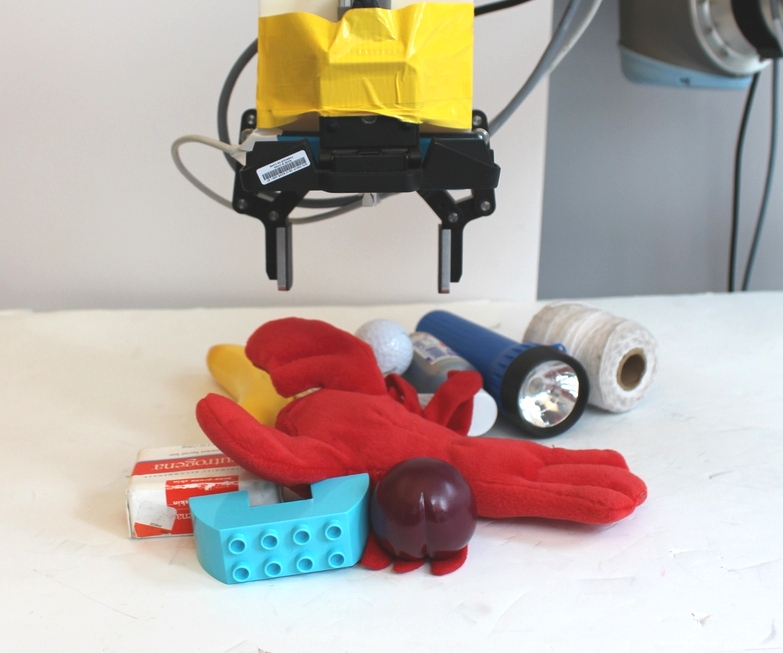}}
    \subfigure{\includegraphics[width=\mywidth]{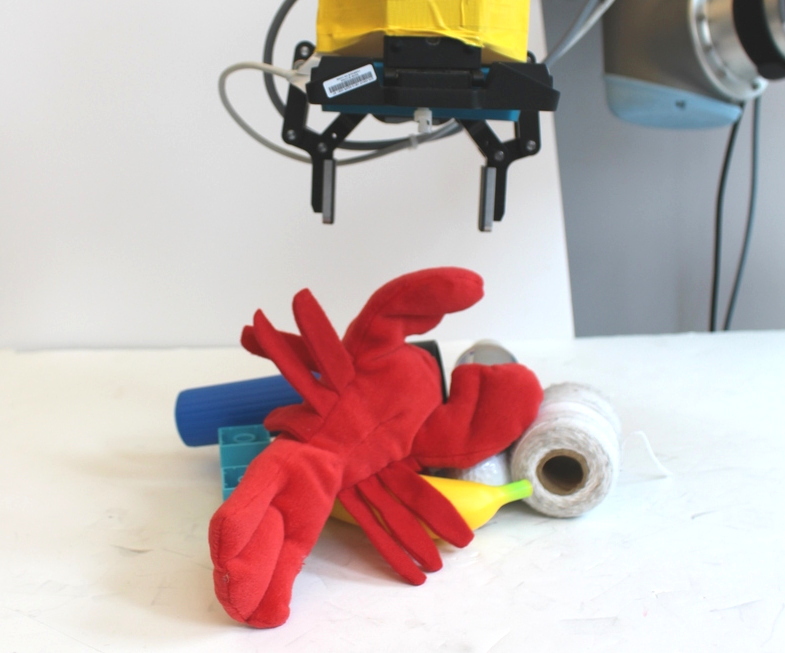}}
    \subfigure{\includegraphics[width=\mywidth]{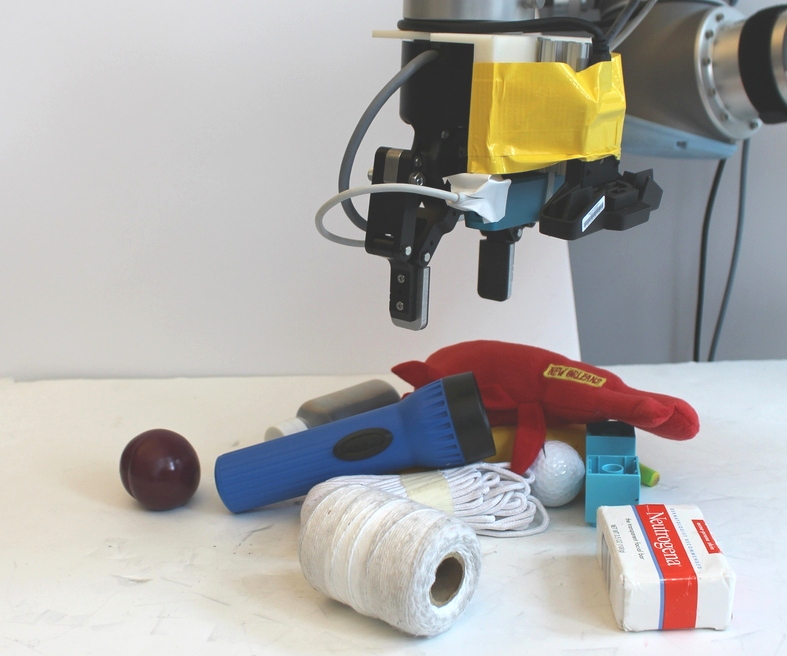}}
    \subfigure{\includegraphics[width=\mywidth]{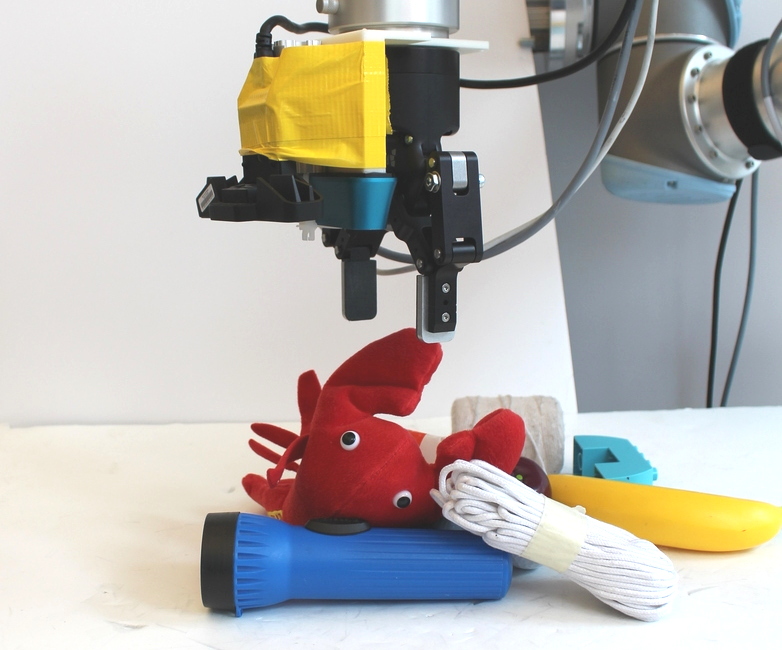}}
    \subfigure{\includegraphics[width=\mywidth]{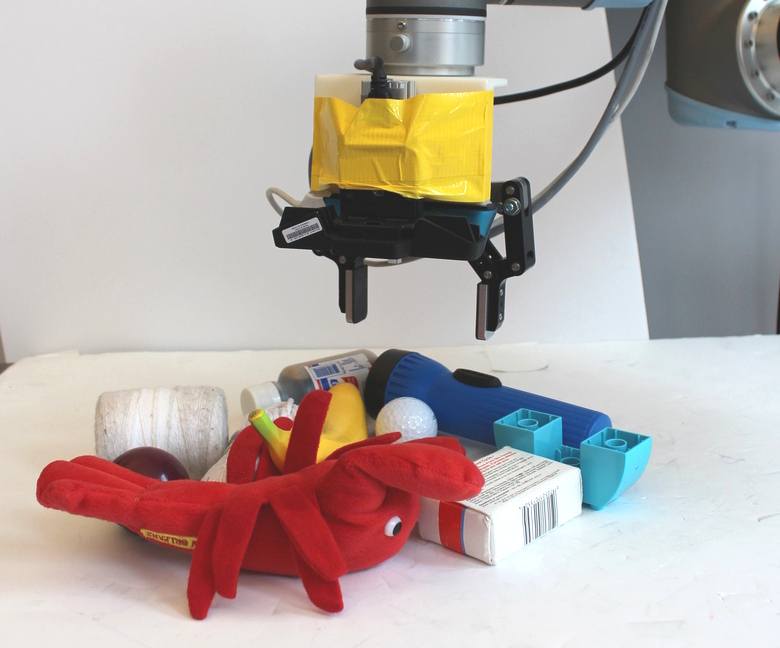}}\\\vspace{-0.12in}    
    \subfigure{\includegraphics[width=\mywidth]{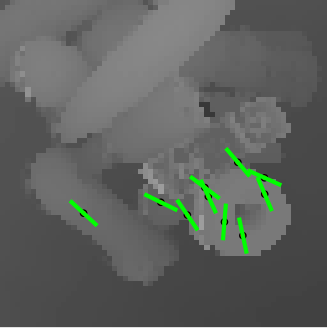}}    
    \subfigure{\includegraphics[width=\mywidth]{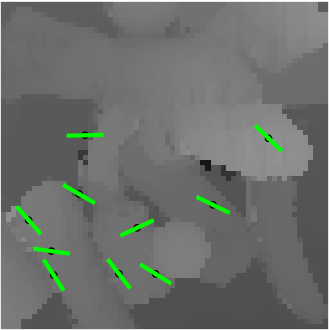}} 
    \subfigure{\includegraphics[width=\mywidth]{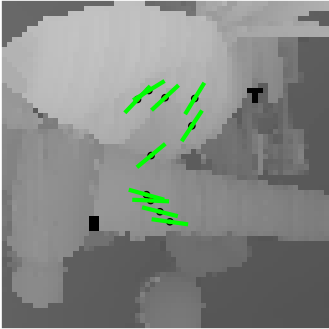}}    
    \subfigure{\includegraphics[width=\mywidth]{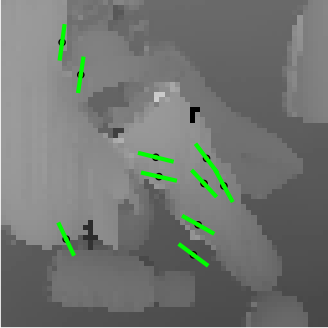}} 
    \subfigure{\includegraphics[width=\mywidth]{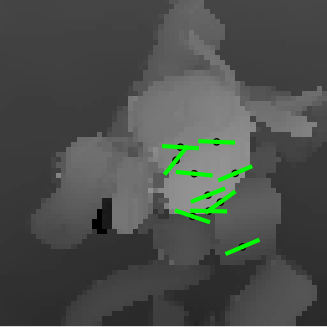}}    
    \subfigure{\includegraphics[width=\mywidth]{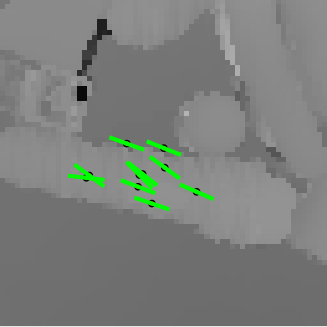}}      
    \caption{Example of cluttered scenes with corresponding depth images and grasps with predicted low distance-to-nearest-grasp.}
    \label{gripper_with_clutter_6x}
\end{figure*}

\subsection{Grasping objects in dense clutter}
We also evaluate our grasp controller in dense clutter. Here, we put the ten objects from our test set into a box, shake the box to mix up the objects, and empty it onto the table in front of the robot. An example of such a scenario is shown in Figure~\ref{gripper_with_clutter_6x}. A run is terminated when three consecutive executions result in grasp failures or when the remaining objects are out of the sensor's view. In total, we perform ten runs of this experiment.

The robot attempted 74 grasps using our controller over the ten runs of this experiment. Out of those 74 attempts, eight are failures (88.9\% grasp success rate). Five out of the eight failures are caused by the object slipping out of the fingers during the grasp, two are caused by slight misplacements in the final pose, and one is caused by a collision between a finger and the object which moved the object out of the hand's closing region. In comparison, the GPD method attempted 96 grasps over ten runs of this experiment. Only five out of those 96 attempts were not successful (94.8\% grasp success rate). Out of the five failures, two are perceptual errors (invalid or missing sensor data) and three are caused by the object slipping out of the hand while closing the fingers. While GPD achieves a higher grasp success rate than our controller in this scenario, we think that the controller would achieve a similar performance if it were given more iterations and the correction movement during each iteration was smaller (such as we did for the simulation).

\subsection{Grasping objects with changing orientations}

This experiment evaluates the performance of our controller versus the GPD baseline for a dynamic scenario where the human experimenter manually shifts the positions of the objects once during each grasp trial. To accomplish this, we pour the pile of cluttered objects onto a piece of paper and then shift the paper by a random amount after the third controller iteration. Over the ten runs of this experiment, the robot attempted 75 grasps using our controller. 17 out of those 75 attempts were failures (77.3\% grasp success rate). In comparison, GPD only attempted 49 grasps, out of which 38 were failures (22.5\% grasp success rate). The better performance of our controller makes sense because it is able to react to the shift whereas GPD cannot: it simply proceeds to execute the grasp as if the object pile had not been shifted. This is a general advantage of a closed-loop controller relative to typical grasp perception methods~\cite{Pinto2016,Redmon2015,Lenz2015,Kappler2015,Gualtieri2016}.

%%%%%%%%%%%%%%%%%%%%%%%%%%%%%%%%%%%%%%%%%%%%%%%%%%%%%%%%%%%%
\section{Discussion}
\vspace{-0.05in}
We developed a visuomotor controller that uses visual feedback from the depth sensor mounted on the gripper to dynamically correct for misalignment with the object during grasping. We trained a deep CNN model with simulated sensor data that directly learns the distance function for a given depth image and grasp pose action. Generation of training data in simulation was more efficient than generation on a real robot. We found that the CNN model trained with simulated depth images transfers well to the domain of real sensor images after processing images to correct invalid depth readings.

Our controller was able to react to shifting objects and to inaccurate placement of the gripper relative to the object to be grasped. In simulation experiments, our approach compensated for significant noisy kinematics while a one-shot GPD baseline did not. Moreover, our controller using the CNN model corrected for perceptual errors present in one-shot prediction. 
Real world experiments demonstrated that our method also works well on the robot with noisy sensor images. Our performance was comparable to the GPD baseline. We expect to improve our controller further, such that we can execute more and smaller corrections while moving the gripper faster during grasping. Results in simulation showed that, by using a controller with fine-grained adjustments, we can exceed the performance of the GPD baseline, especially in the presence of kinematic noise. The UR5 robot used in our experiments has fairly precise forward kinematics. For experiments on a robot with more noisy kinematics (e.g., the Baxter robot) we expect to see a significant advantage of our method.

\clearpage

\acknowledgments{This work has been supported in part by the National Science Foundation through IIS-1427081, CCF-1723379, IIS-1724191, and IIS-1724257, NASA through NNX16AC48A and NNX13AQ85G, and ONR through N000141410047.}

%===============================================================================

\bibliography{main}  % .bib

\end{document}